%% file: main.tex
\documentclass{article}

\usepackage{iclr2027_conference,times}
\usepackage{float}
\input{math_commands.tex}

\input{imports}

\input{title}

\input{authors}

\iclrfinalcopy

\begin{document}

\maketitle
\lhead{Preprint. Under review.}
\equalcontribnote

\input{sections/abstract}
\input{sections/introduction}
\input{sections/related_work}
\input{sections/prelims}
\input{sections/methodology}
\input{sections/experiments}
\input{sections/results}
\input{sections/conclusion}
\input{sections/mandatory_ICLR_sections}

\bibliography{references}
\bibliographystyle{iclr2027_conference}

\appendix
\input{sections/appendix}

\end{document}

%% file: math_commands.tex
\usepackage{amsmath,amsfonts,bm}

\def\eqref#1{equation~\ref{#1}}

\def\1{\bm{1}}

\DeclareMathAlphabet{\mathsfit}{\encodingdefault}{\sfdefault}{m}{sl}
\SetMathAlphabet{\mathsfit}{bold}{\encodingdefault}{\sfdefault}{bx}{n}

%% file: imports.tex
\usepackage{hyperref}
\usepackage{url}
\usepackage{graphicx}
\usepackage{booktabs}
\usepackage{xcolor}
\hypersetup{
  pdftitle={Particle Belief Representations for Reinforcement Learning under Partial Observability},
  pdfauthor={Himanshu Gupta, Brendan Crowe, Christoffer Heckman, Bradley Hayes, Zachary Sunberg}
}

%% file: title.tex
\title{Particle Belief Representations \texorpdfstring{\\}{ }for Reinforcement Learning \texorpdfstring{\\}{ }under Partial Observability}

%% file: authors.tex
\author{Himanshu Gupta$^{1\dagger}$, Brendan Crowe$^{2\dagger}$, Christoffer Heckman$^{2}$, \\
\textbf{Bradley Hayes$^{2}$, Zachary Sunberg$^{1}$} \\
$^{1}$Department of Aerospace Engineering Sciences, University of Colorado Boulder \\
$^{2}$Department of Computer Science, University of Colorado Boulder \\
Email: \texttt{\{firstname.lastname\}@colorado.edu}
}

\newcommand{\equalcontribnote}{%
  \begingroup
  \renewcommand{\thefootnote}{}%
  \footnotetext{$^{\dagger}$These authors contributed equally to this work.}%
  \endgroup}

%% file: sections/abstract.tex
\begin{abstract}

Particle filters flexibly represent complex, multimodal beliefs but expose them as
weighted, unordered sets incompatible with conventional fixed-dimensional policies.
We study particle-belief encodings as inputs to reinforcement-learning policies while holding belief estimation and policy optimization fixed. We compare statistical
summaries, cumulant-generating-function (CGF) features, and permutation-invariant
set encoders. Trainable encoders are learned end-to-end or pretrained through
belief reconstruction or task-specific prediction. Across eight tasks, pooling-based
set encoders and gradient CGF features are reliable end-to-end, while the Set
Transformer achieves the highest performance with appropriate pretraining.
Task-specific pretraining helps when its target captures decision-relevant belief
structure, whereas reconstruction pretraining and latent alignment provide more selective benefits.
For selecting and training particle-belief encoders in partially observable Markov decision processes, our results show that an encoding that reconstructs the belief well is not necessarily a good input for control.

\end{abstract}

%% file: sections/introduction.tex
\section{Introduction} \label{sec_introduction}

Partial observability is a critical challenge in reinforcement learning (RL).
Most foundational RL algorithms were developed for Markov decision processes (MDPs), where the current observation contains sufficient information to act optimally.
However, an agent rarely has perfect access to the state and the partially observable Markov decision process (POMDP) is a more appropriate model.
Unfortunately, in a POMDP, crucial information for choosing an optimal action could be contained in \emph{any} observation in the history, not just the current one.
A helpful classical result is that the POMDP \emph{belief}, the state distribution given the observation history, is a sufficient statistic for decision making~\citep{witness_kaelbling}.
In mainstream RL research, beliefs are rarely explicitly estimated, with high profile research projects instead using an \emph{implicit} summary of the observation history through recurrent neural networks~\citep{vinyals2019grandmaster} or more recently tokenization and world-model-based prediction~\citep{hafner2025training}.
In contrast, in aerospace and robotics it is routine to calculate beliefs over unknown states before applying control actions~\citep{lauri2022partially}.

Though belief estimation has downsides including the potential to track irrelevant information, it has two significant advantages.
The first is \emph{composability}.
Beliefs provide a natural and unambiguous interface to communicate state estimates and uncertainty.
For instance, if a policy can interpret beliefs, then sensors such as cameras, lidars, and inertial measurement units, and even the dynamics models or belief-estimation algorithms themselves, can be developed and improved independently without accessing or changing the policy.
This composability framework could enable broader re-use and proliferation of AI components provided by a large ecosystem of vendors rather than only monolithic end-to-end trained systems~\citep{deglurkar2023compositional}.
The second advantage is \emph{interpretability}.
In the event of a system failure, investigators can diagnose whether the belief was inconsistent with the true state or the policy made a poor decision due to misspecified objectives or another issue.
Moreover, properties of the belief, such as the level of uncertainty or the probability of a dangerous state, can be directly constrained or penalized in the reward function.

\begin{figure}[t]
  \centering
  \includegraphics[width=\textwidth]{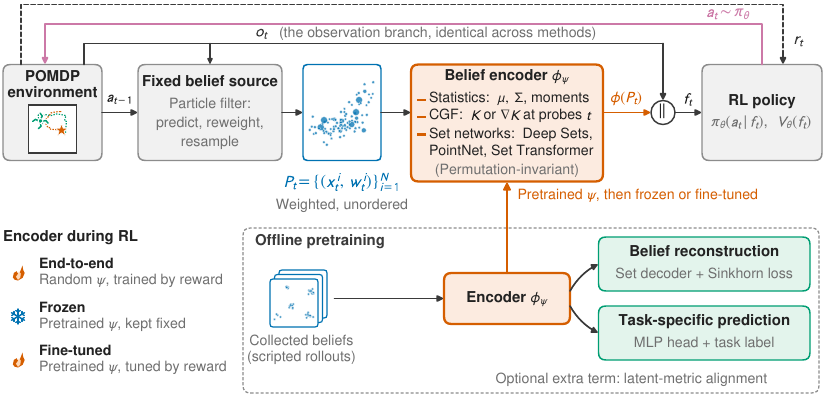}
  \caption{System overview. A fixed particle filter produces a weighted
    particle belief, which a belief encoder maps to a fixed-dimensional code for the RL policy. We hold the belief source and downstream policy
    optimization fixed while varying the encoder and, optionally, its offline
    pretraining and transfer mode.}
    \label{fig:system-overview}
\end{figure}

In all but the simplest problems, however, maintaining an exact belief is computationally intractable. While parametric methods such as Kalman filters generally impose a unimodal Gaussian approximation, particle filters can approximate complex, multimodal belief distributions and easily accommodate nonlinear dynamics.
The belief is a weighted collection of state-estimate ``particles'':
\begin{equation}\label{eq:pf}
    b_t^N
    =
    \left\{
        \left(x_t^{i}, w_t^{i}\right)
    \right\}_{i=1}^{N},
    \qquad
    \sum_{i=1}^{N} w_t^{i} = 1,
    \qquad  w_t^i \ge 0
\end{equation}
Particle beliefs also support Monte Carlo estimation of belief-dependent quantities, such as expected rewards, without constructing an explicit discretization of the full state space~\citep{sparse_pft}.
This belief approximation is broadly applicable, but it creates a fundamental representation problem.
A particle belief is not naturally represented as a fixed-dimensional vector.
It is an unordered, potentially variable-size collection whose elements jointly approximate a probability distribution.
Consequently, a useful particle-belief representation should be insensitive to particle ordering, account for nonuniform particle weights, preserve decision-relevant distributional structure, and remain robust to the Monte Carlo variability of the filter.
Standard neural-network inputs satisfy few of these requirements.
Flattening the particles imposes an arbitrary ordering and ties the policy architecture to a fixed particle count.
Sorting the particles introduces a hand-designed ordering that may be discontinuous and task-dependent.
Compressing the belief to its mean and covariance is computationally efficient~\citep{betazero}, but can make fundamentally different beliefs indistinguishable.
For example, a unimodal belief and a symmetric multimodal belief can have similar low-order statistics while requiring entirely different information-gathering or risk-sensitive actions.
The choice of belief encoding is not a mere implementation detail; it determines which beliefs can be discriminated, what policy paradigms can ingest it and thus the class of problems it can solve.

Despite the importance of this design choice in contexts where the composability, interpretability, and flexibility advantages of particle filters are important, particle-belief representation has rarely been studied as an independent problem.
Existing work has generally studied particle-belief representations within broader architectures~\citep{ma2020dpfrl,solinas2025nbf,abdulsamad2025smc}.
For example, \citet{ma2020dpfrl} introduce a moment-generating function (MGF) representation as part of an end-to-end particle-filter reinforcement-learning system, making it difficult to isolate the effect of the representation itself.
This leaves a central question unresolved:
\begin{quote}
\emph{What are the most effective particle-belief encodings for reinforcement learning, and how should they be constructed?}
\end{quote}

\noindent\textbf{Contributions:} We study this question in particle-filter belief MDPs~(Figure~\ref{fig:system-overview}).
We perform a controlled comparison of particle-belief encoders (low-order weighted statistics, cumulant-generating-function (CGF) features, permutation-invariant pooling networks, and attention-based set encoders) under an identical particle filter, RL algorithm, and policy architecture, thereby isolating the effect of the belief representation.
Each trainable encoder is evaluated under end-to-end training and under reconstruction or task-specific pretraining, with frozen or fine-tuned transfer and optional latent-metric alignment.
The eight tasks combine established POMDP domains with diagnostic tasks that isolate active information gathering, multimodality, cluster comparison, and parity-constrained belief structure.
Our results show that an encoding that reconstructs the belief well is not necessarily a good input for control. Pooling-based set encoders and gradient CGF features are reliable under end-to-end training, whereas the Set Transformer achieves the highest performance with appropriate pretraining. Task-specific pretraining produces large gains when its target captures decision-relevant belief structure, but can be harmful when it discards uncertainty needed for information gathering; reconstruction pretraining and latent alignment provide more selective benefits.

%% file: sections/related_work.tex
\section{Related Work}
\label{sec:related}

\textbf{POMDPs:} \label{sec:related:pomdps}
Since solving a POMDP exactly is intractable~\citep{pomdp_complexity_papadimitriou_1987}, numerous methods approximate its solution using online tree search~\citep{despot,pomcpow}, which has been applied to problems such as navigation and target interception in crowds~\citep{gupta_ped_navigation_holonomic,gupta2026beyond}.
These methods simulate future action--observation trajectories from the current belief.
Although effective, they repeat the search at every decision and are constrained by search depth, branching, and the cost of belief updates.
Extensions of online tree search support continuous state, action, and observation spaces~\citep{vomcpow}, but efficiently exploring continuous actions and reasoning over long horizons remain challenging.
BetaZero addresses the latter by learning policy and value approximations that guide and bootstrap tree search for discrete action spaces~\citep{betazero}.
Like our approach, BetaZero maintains a particle belief, but its neural networks receive only a Gaussian-style summary consisting of the particle mean and standard deviation.
BETS instead gives these networks a learned set embedding of the belief, propagated by a neural Bayesian filter in place of the particle filter~\citep{solinas2026bets}.
Direct policy learning provides a complementary alternative by amortizing decision making into training and avoiding online tree search at deployment.
This makes the choice of particle-belief encoding especially important, since the policy can act only on information preserved by the representation.

\noindent\textbf{Reinforcement Learning for POMDPs:}
Many reinforcement-learning approaches handle partial observability without explicitly estimating a belief, instead conditioning the policy on a recent observation window or a learned representation of the interaction history.
Frame stacking provides a fixed history window~\citep{mnih2015human}, while recurrent policies~\citep{bakker2001lstm,wierstra2007recurrent,deepq} and transformers~\citep{parisotto2020gtrxl} learn implicit summaries over longer histories.
Although effective, these approaches couple representation learning to policy optimization and do not provide an explicit, reusable representation of uncertainty.
Other work incorporates belief structure directly through differentiable filters~\citep{karkus2017qmdpnet,jonschkowski2018differentiable,karkus2018pfnet} and particle-based latent states~\citep{ma2020pfrnn,DVRL}.
Closest to our setting, DPFRL jointly learns a particle filter and summarizes its belief for the policy using the particle mean and learned MGF features~\citep{ma2020dpfrl}, and \citet{abdulsamad2025smc} encode a fixed particle filter's belief with a Set Transformer in a new policy-optimization method.
In these methods and in BETS, the encoding comes with a new filter, planner, or learning algorithm, so its contribution is not isolated.
We instead hold the particle filter and reinforcement-learning procedure fixed while varying the particle-belief encoder and how it is trained.

\noindent\textbf{Moment- and Cumulant-Generating Functions:}
The MGF characterizes a distribution when it exists in a neighborhood of the origin, while derivatives of its logarithm, the CGF, recover the distribution's cumulants~\citep{mccullagh1987tensor,billingsley1995probability}.
Although the complete MGF, or equivalently the complete CGF, uniquely characterizes the distribution under these conditions, evaluating either function at finitely many probe points produces only a finite-dimensional summary, a special case of kernel mean embeddings~\citep{muandet2017kernel}.
DPFRL uses MGF features evaluated at learned points to summarize a learned particle belief for reinforcement learning~\citep{ma2020dpfrl}.
We use the CGF because its logarithmic form retains the information in the MGF while enabling a numerically stable log-sum-exp computation.
We evaluate the empirical CGF and also its gradient with respect to the probe locations, which DPFRL does not use.

\noindent\textbf{Permutation-Invariant Set Encoders:}
Particle beliefs are unordered sets, making permutation invariance a natural requirement for their encoders.
Deep Sets processes elements independently before sum pooling, while PointNet uses symmetric max pooling~\citep{zaheer2017deep,pointnet}; neither explicitly models interactions among elements before aggregation.
The Set Transformer instead uses permutation-equivariant self-attention to model such interactions before attention-based pooling, with inducing points reducing the cost of attention~\citep{vaswani2017attention,lee2019settransformer}.
Related set architectures have been used in RL to process unordered sensory inputs and relational entity representations~\citep{tang2021sensory,zambaldi2019relational}.
Set autoencoders can be pretrained by reconstructing unordered inputs using Earth Mover's Distance (EMD), which measures the minimum cost of transporting mass between two distributions~\citep{fan2017point,achlioptas2018learning,rubner2000emd}.
We use a differentiable Sinkhorn-based optimal-transport loss  to reconstruct weighted particle beliefs~\citep{sinkhorn}.
Generative distribution embeddings~\citep{fishman2025gde} replace the set decoder with a conditional generative model, and \citet{solinas2025nbf} use such embeddings as the belief state of a filter. Both assess how accurately the distribution or the hidden state is recovered; we instead assess the performance of a policy trained on the encoding.
In RL from images, representation learning can be decoupled from policy learning by pretraining the encoder and using it frozen in the agent~\citep{stooke2021atc}; we follow this route for particle-belief encoders, comparing frozen and fine-tuned transfer.
Self-supervised losses can instead be trained jointly with the RL objective to improve sample efficiency, as in latent self-prediction~\citep{schwarzer2021spr} and temporal contrastive learning~\citep{zheng2023taco}; we do not study such objectives.

%% file: sections/prelims.tex
\section{Preliminaries}
\label{sec:prelim}

\textbf{POMDPs:}
A POMDP is defined by a tuple $\mathcal{M}=(\mathcal{S},\mathcal{A},\mathcal{O},T,Z,R,\gamma,b_0)$, where $\mathcal{S}$, $\mathcal{A}$, and $\mathcal{O}$ denote the state, action, and observation spaces, which may be continuous, discrete, or hybrid. $T(s'\mid s,a)$ is the transition model; $Z(o\mid s',a)$ is the observation model; $R(s,a)$ is the reward; $\gamma$ is the discount factor; and $b_0$ is the initial-state distribution. At step $t$, the agent takes action $a_t$, the state transitions to $s_{t+1}\sim T(\cdot\mid s_t,a_t)$, and the agent receives reward $r_t=R(s_t,a_t)$ and observation $o_{t+1}\sim Z(\cdot\mid s_{t+1},a_t)$.
Because the state is not directly observed, the agent maintains a belief $b_t$, a probability distribution over $\mathcal{S}$. After action $a_t$ and observation $o_{t+1}$, Bayes' rule gives $b_{t+1}(s')\propto Z(o_{t+1}\mid s',a_t)\int_{\mathcal{S}}T(s'\mid s,a_t)b_t(s)\,ds$.
Treating the belief as the state yields a \emph{belief MDP}. With reward $\rho(b,a)=\mathbb{E}_{s\sim b}[R(s,a)]$, its optimal policies are also optimal for the original POMDP~\citep{witness_kaelbling}. We seek $\pi^{\ast}(a\mid b)$ maximizing $\mathbb{E}_{\pi}[\sum_t\gamma^t r_t]$.

\noindent\textbf{Particle filters:}
A particle filter~\citep{particle_filter} approximates the belief by the normalized weighted set $b_t^N$ in~\eqref{eq:pf}. Each update propagates particles through the transition model, reweights them using the latest observation, and optionally resamples when particle degeneracy warrants it. Our framework is agnostic to the particular filtering algorithm and requires only a weighted particle set of the form in~\eqref{eq:pf}. Belief encoders consume the particle--weight pairs as an unordered set and must therefore be invariant to their joint permutation (Section~\ref{sec:methods:encoders}).

\noindent\textbf{Problem statement:}
We decompose the state as $s_t=(s_t^{\mathrm{o}},s_t^{\mathrm{h}})$, where $s_t^{\mathrm{o}}$ is observed and $s_t^{\mathrm{h}}$ is hidden. At time $t$, the filter supplies a weighted particle approximation $b_t^N$ to the posterior over $s_t^{\mathrm{h}}$. Together, $s_t^{\mathrm{o}}$ and this posterior specify the complete belief over $s_t$, which is sufficient for control.
A belief encoder maps the particle belief to a fixed-dimensional representation, $\phi(b_t^N)\in\mathbb{R}^{d_\ell}$. Because the ordering of the particle--weight pairs is arbitrary, reordering these pairs cannot change $\phi(b_t^N)$. The policy and value networks receive $f_t=[s_t^{\mathrm{o}}\Vert\phi(b_t^N)]$, where $\Vert$ denotes concatenation.

%% file: sections/methodology.tex
\section{Methodology}
\label{sec:methods}
A particle belief can be mapped to a fixed-dimensional, permutation-invariant representation in many ways, but different representations preserve different aspects of the belief. It remains unclear which information is important for downstream control and, for learned encoders, what supervision best preserves it. Training only through policy gradients provides sparse and indirect supervision; task-agnostic reconstruction may retain irrelevant variation, while task-specific prediction may discard uncertainty needed for information gathering. We therefore study particle-belief representation along three coupled axes: encoder architecture, pretraining objective, and transfer to reinforcement learning~(Figure~\ref{fig:system-overview}). For each trainable encoder, we compare end-to-end policy learning with reconstruction- and task-specific pretraining, followed by either frozen or fine-tuned transfer; we additionally study whether aligning latent and belief-space geometry improves transfer. Throughout, we hold the belief estimate, observation branch, policy and value architectures, and RL algorithm fixed. This design reveals which representations work and how training signals affect performance.
\subsection{Belief encoders}
\label{sec:methods:encoders}

\textbf{Statistical summaries:}
The Gaussian encoder concatenates the particle belief's weighted mean and covariance. The moment encoder concatenates its weighted mean with the coordinate-wise central moments $\mu_r=\mathbb{E}[(X-\mathbb{E}[X])^r]$ for $r=2,\ldots,k$. Both are inexpensive permutation-invariant encoders but discard structure beyond these statistics.

\noindent\textbf{Cumulant-generating features:}
For a random vector $X$, the cumulant-generating function (CGF) is
$K(t)=\log\mathbb{E}[e^{t^\top X}]$ when the expectation is finite. Its
derivatives at the origin recover the cumulants of $X$, including its mean and
covariance. Evaluating the CGF away from the origin therefore provides nonlinear
features that can capture distributional structure beyond a fixed collection of
low-order moments. For a weighted particle belief $\{(x_i,w_i)\}_{i=1}^N$, the
empirical CGF and its gradient are
\par\noindent
\begin{minipage}[t]{0.39\linewidth}
\begin{equation}
\hat K(t_j)=\log\!\sum_{i=1}^N w_i e^{t_j^\top x_i}.
\end{equation}
\end{minipage}\hfill
\begin{minipage}[t]{0.59\linewidth}
\begin{equation}
\nabla \hat K(t_j)=\sum_{i=1}^N
\frac{w_i e^{t_j^\top x_i}}{\sum_{\ell}w_{\ell}e^{t_j^\top x_{\ell}}}x_i.
\end{equation}
\end{minipage}
\par\vspace{\belowdisplayskip}\noindent
The gradient is the particle mean under exponentially tilted weights and
emphasizes particles aligned with the probe direction $t_j$. Given probes
$\{t_j\}_{j=1}^m$, we form the representation by concatenating either the scalar
values $\hat K(t_j)$ or the vectors $\nabla\hat K(t_j)$. Probe locations are
initialized over a domain-scaled range and are either held fixed or treated as
trainable parameters.

\noindent\textbf{Learned set networks:}
Deep Sets and our PointNet-style encoder augment each particle state with
its scaled normalized weight, $[x_i,Nw_i]$, and process every particle using a shared MLP.
Deep Sets aggregates the resulting features using the weighted sum
$\sum_i w_i\psi([x_i,Nw_i])$, whereas PointNet uses coordinate-wise max pooling
after excluding particles with zero weight. Positive weights therefore enter
PointNet through the per-particle weight channel rather than through the pooling
operation. A second MLP maps the pooled features to the belief representation.
The two encoders otherwise use matched architectures, isolating weighted-sum
versus max aggregation. The Set Transformer receives the same augmented particle
input and uses induced self-attention to model interactions among particles
before attention-based pooling~\citep{lee2019settransformer}.

\subsection{Offline pretraining}
\label{sec:methods:pretraining}

We pretrain only parameterized belief encoders using two objectives; in both cases, we discard the auxiliary decoder or prediction head and transfer only the encoder to reinforcement learning.

\noindent\textbf{Belief reconstruction:}
We jointly train the encoder and a shared set decoder to reconstruct the weighted empirical belief. The decoder uses $N$ learned queries to attend to the latent representation and maps the resulting features through an MLP to $N$ uniformly weighted particle locations. A natural reconstruction loss is Earth Mover's Distance (EMD), which measures the minimum cost of transporting probability mass from the input belief to its reconstruction~\citep{rubner2000emd}. Computing exact EMD is expensive, so we instead minimize the debiased Sinkhorn divergence, an efficient and differentiable approximation~\citep{sinkhorn,feydy2019interpolating}.

\noindent\textbf{Task-specific prediction:}
Alternatively, an MLP prediction head reads the encoded belief and predicts a task-relevant label. This objective tests whether retaining information directly relevant to control provides a better initialization than reconstructing the complete belief.

\noindent\textbf{Latent metric alignment:}
As an optional regularizer for either pretraining objective, we encourage distances between latent representations to preserve distances between beliefs. For a minibatch, let $d_z$ contain the cosine distances between sampled pairs of latent codes and $d_{\mathcal{B}}$ the corresponding Sinkhorn distances between their particle beliefs, computed on the fly. We add $\mathcal{L}_{\mathrm{align}}=\lambda[1-\mathrm{corr}(d_z,d_{\mathcal{B}})]$ to the pretraining loss after a short warm-up. Correlation makes the regularizer insensitive to the relative scale of the two metrics.

\subsection{Encoder training and transfer}
\label{sec:methods:transfer}

We evaluate three strategies for trainable encoders: (i) \emph{end-to-end}, which jointly optimizes a randomly initialized encoder with the policy and value networks from reward alone; (ii) \emph{frozen transfer}, which initializes the encoder from pretraining and updates only the policy and value networks; and (iii) \emph{fine-tuned transfer}, which also updates the encoder at $0.1$ times the RL learning rate.

%% file: sections/experiments.tex
\section{Experiments}
\label{sec:experiments}

Our experiments evaluate the three axes of Section~\ref{sec:methods}: which encoder inductive biases suit beliefs with qualitatively different geometry, whether offline pretraining improves control, and whether a pretrained encoder should be frozen or fine-tuned.
All conditions share a proximal policy optimization (PPO) harness~\citep{PPO} and differ only in the belief encoder and its initialization.

\subsection{Task suite}
\label{sec:experiments:tasks}

\begin{figure}[t]
  \centering
  \includegraphics[width=\textwidth]{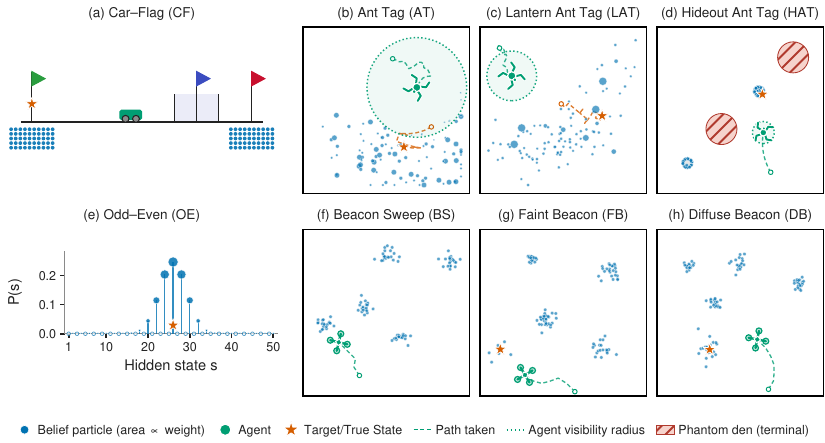}
  \caption{The eight evaluation tasks and their particle beliefs.
    Marker area is proportional to particle weight.}
    \label{fig:task-suite}
\end{figure}

We evaluate eight tasks from four environment families
(Figure~\ref{fig:task-suite}) and report them in three groups that differ in
what the policy must extract from the belief: whether it has to move to gather
information, compare the components of a multimodal belief, or resolve a
categorical ambiguity.

\noindent\textbf{Active information gathering (CF, AT, LAT, HAT):}
The policy must act to reduce its own uncertainty.
We adapt Car--Flag and the Ant Tag family from the robot POMDP domains of~\citet{nguyen2021penvs}.
Car--Flag is a simple control task in which a hidden bit selects the rewarding
end of a one-dimensional track and is revealed in an intermediate information
region, yielding three possible beliefs.
The three Ant Tag variants require a MuJoCo ant to locate and tag a moving
target whose unknown position is tracked using a weighted particle filter (Appendix~\ref{app:anttag-details}). In
Ant Tag, the target becomes observable within a sensing radius and evades the ant
more aggressively at close range. Lantern Ant Tag uses the same dynamics but a
smaller sensing radius, leaving the target unobserved for most of the episode
and increasing reliance on the belief.
Hideout Ant Tag is a deliberately moment-matched variant. At the start of each
episode, one of two den configurations is active. Each configuration contains
two dens, and the configurations are point reflections of each other about the
arena center. The target is in one of the active configuration's dens. Within
each configuration, a high-probability den near the center counterbalances a
low-probability den farther away, placing the belief mean at the center.
The configurations therefore have identical means and covariances despite
having different multimodal particle distributions. The inactive pair acts as
terminal hazards, so a policy that uses only the mean and covariance cannot identify
the safe configuration, whereas success requires retaining higher-order
structure in the belief.

\noindent\textbf{Beacon tasks (BS, FB, DB):}
We introduce three new search-and-rescue environments that share a $20\times20$
arena, velocity actions in $[-1,1]^2$, and 60-step episodes (Appendix~\ref{app:beacon-details}). Five survivors carry
radio beacons at hidden, well-separated locations $\mu_j$. At each step, beacon
$j$ transmits with probability $\rho_j$, and each received signal provides an
anonymous location estimate drawn from $\mathcal{N}(\mu_j,\sigma_j^2 I)$. We
represent the long-run distribution of these signals as the mixture
$p(x)\propto\sum_j \rho_j\,\mathcal{N}(\mu_j,\sigma_j^2 I)$ and provide the
policy with a weighted particle representation of this distribution at every
step. This construction isolates belief representation from estimation error.
Beacon Sweep rewards reaching all five survivors, removing each reached
survivor from the belief. Faint and Diffuse Beacon permit only one
rescue: the episode ends when the agent reaches its first survivor. In Faint
Beacon, the buried survivor transmits less frequently and therefore corresponds
to the component with the least mass. In Diffuse Beacon, it produces noisier
location estimates and therefore corresponds to the component with the greatest
spread. The agent is rewarded if it reaches the buried survivor and penalized if
it reaches another survivor or the episode times out. The three tasks therefore
isolate coverage, mass comparison, and scale comparison on otherwise identical
state and action spaces.

\noindent\textbf{Odd--Even (OE):}
A hidden integer $s^\star\in\{1,\ldots,50\}$ is fixed for each 30-step
episode. Each observation is drawn from a Gaussian centered at $s^\star$ but
restricted to integers with the same parity, producing a weighted, comb-shaped
posterior over the candidate integers. We represent this posterior exactly using
one weighted particle per candidate integer (Appendix~\ref{app:oddeven-details}). At each step, the policy predicts
$s^\star$ and receives
reward only for an exact match. Mean and variance capture the posterior's
location and spread but not its parity-constrained, comb-shaped support, so a
Gaussian encoding can conflate beliefs that place mass on different candidate
integers.

\subsection{Experimental comparisons}
\label{sec:experiments:comparisons}

\textbf{End-to-end representations:}
Our end-to-end study compares ten ways of constructing the policy input: the
nine belief encoders defined in Section~\ref{sec:methods:encoders} and a
frame-stacking baseline that concatenates the most recent $k$ raw observations,
thereby treating the finite observation history as an approximate Markov state.
In the Ant Tag variants, each stacked frame contains the target sighting when in range (plus
the filter's den prior in HAT); Odd--Even stacks the observation and time
index; and the beacon tasks stack the current pings. 
We set $k$ to 8 for the Ant Tag variants and to the episode length for Car--Flag (160), Odd--Even (30), and the beacon tasks (60).
On the beacon tasks, the encoders receive the converged signal distribution from the first step, whereas frame stacking starts each episode from an empty history (Appendix~\ref{app:framestack}).
The learned set encoders produce 64-dimensional belief representations and have comparable parameter counts (Appendix~\ref{app:encoders}). For the CGF encoders, we use 64
probe locations, yielding 64 features for $K$ and 128 features for $\nabla K$,
with probe locations either fixed or learned (Section~\ref{sec:methods:encoders}).
The analytic encoders retain their natural dimensionality: the Gaussian encoder outputs the belief mean and covariance,
while the moment encoder uses moments through order eight.
We also tested sensitivity to the number of CGF probes and the maximum moment
order on Faint and Diffuse Beacon; the results are in Appendix~\ref{app:ablation}.

\noindent\textbf{Pretraining and transfer:}
For the three learned set encoders and the two CGF feature maps with learned probes
($K$ and $\nabla K$), we evaluate reconstruction and task-specific pretraining (Appendix~\ref{app:pretraining}).
Each pretrained encoder is transferred to PPO either frozen or fine-tuned, as defined
in Sections~\ref{sec:methods:pretraining} and~\ref{sec:methods:transfer}. The
task-specific targets are the categorical posterior in Odd--Even, the target-cluster
offset in the three beacon tasks, target position in Ant Tag and Lantern Ant Tag,
and posterior mass over candidate dens in Hideout Ant Tag. For each encoder--objective
pair, a single pretrained checkpoint (seed 0) initializes all downstream PPO runs, isolating variation due to downstream policy optimization. We additionally
test whether preserving belief-space geometry improves transfer
by adding the latent metric-alignment term from
Section~\ref{sec:methods:pretraining} to both reconstruction and task-specific
pretraining. We compare each aligned condition with its unaligned counterpart
on all tasks except Car--Flag, which is evaluated only end-to-end, and Hideout
Ant Tag, where unaligned pretraining already reaches the empirical performance
ceiling. We use $\lambda=0.2$ for the Ant Tag variants and
Odd--Even, and $\lambda=0.02$ otherwise, as selected by preliminary
hyperparameter sweeps.

\subsection{Training and evaluation protocol}
\label{sec:experiments:protocol}

For each task, all encoders use the same particle filter, preprocessing, PPO implementation, policy and value architectures, training horizon, and curriculum. We train five seeds per condition for 1M steps on Car--Flag, 3M on Odd--Even and the beacon tasks, and 6M on the Ant Tag tasks. We report mean task success of deterministic policies over 300 episodes, with standard errors across seeds.

%% file: sections/results.tex
\section{Results}
\label{sec:results}

\begin{figure}[t]
  \centering
  \includegraphics[width=\textwidth]{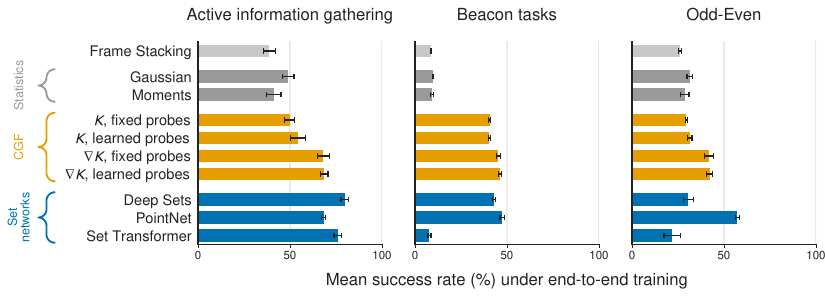}
  \caption{End-to-end training. Mean success rate per task group for each encoder, colored by family. Error bars: standard error of the per-seed group mean over five PPO seeds.}
    \label{fig:results-e2e-groups}
\end{figure}

Figures~\ref{fig:results-e2e-groups} and~\ref{fig:results-pretrain-stacked} summarize the main results, with per-task results reported in Appendix~\ref{app:per-task-results}. Error bars show standard errors across five PPO seeds.

The simple baselines are not uniformly weak. Gaussian statistics, higher-order moments, and frame stacking are competitive on Car--Flag, while moments and frame stacking nearly match the best encoders on Hideout Ant Tag. On the remaining tasks, however, these baselines generally trail the richer particle-belief representations.

On the active-information-gathering tasks, set encoders perform best overall under end-to-end training (left panel of Figure~\ref{fig:results-e2e-groups}), with the gradient CGF representation close behind. Among the CGF encoders, the gradient features we add match or outperform the scalar features, the logarithm of the MGF features of \citet{ma2020dpfrl}, on all eight tasks, e.g., 75\% against 58\% success on Ant Tag (Table~\ref{tab:e2e}, Appendix~\ref{app:per-task-results}). These tasks require not only representing the belief, but also using uncertainty to gather information while controlling the agent. Across the two information-gathering tasks included in Figure~\ref{fig:results-pretrain-stacked}, pretraining generally does not improve upon end-to-end training and sometimes reduces performance. Adding latent alignment often narrows this gap, particularly for reconstruction-pretrained encoders, but does not consistently exceed end-to-end training performance.

The beacon tasks show the clearest effect of the pretraining objective. Under end-to-end training, several encoders nearly solve Beacon Sweep, but all perform poorly on Faint Beacon and Diffuse Beacon; consequently, no encoder exceeds 50\% success when averaged across the three tasks. The Set Transformer is a notable exception to the strong end-to-end performance on Beacon Sweep: it performs poorly when trained end to end, but approaches perfect success after reconstruction pretraining. Because all encoders use identical PPO settings on this task (Appendix~\ref{app:ppo}) and the same architecture succeeds after pretraining, we read this as an optimization difficulty rather than a capacity limit. Thus, learning to reconstruct the particle set appears sufficient to recover the belief geometry needed to sweep the beacons. However, the same pretraining provides little benefit on Faint Beacon and Diffuse Beacon, which require comparing properties of different particle clusters. Task-specific pretraining improves all three set encoders on these comparison-based tasks, but Deep Sets and PointNet still fall short of the Set Transformer. Comparisons between particle clusters may be easier for the Set Transformer because its attention mechanism directly models particle interactions; Deep Sets and PointNet instead encode particles independently before pooling. The CGF representations gain little from pretraining, and latent alignment provides no additional benefit.

\begin{figure}[t]
  \centering
  \includegraphics[width=\textwidth]{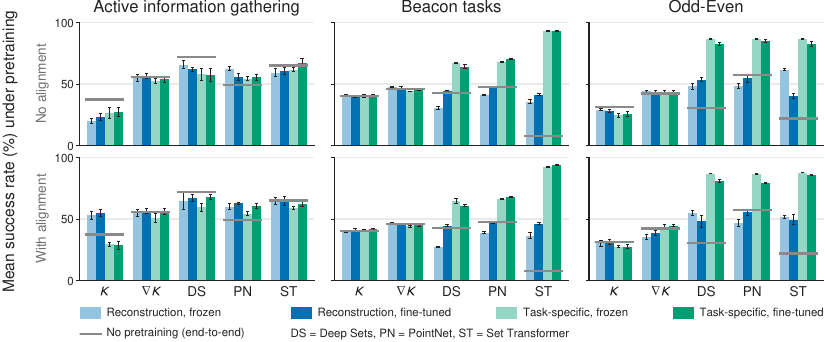}
  \caption{Pretraining and transfer for the five trainable encoders. Top:
  reconstruction and task-specific pretraining, each frozen or fine-tuned;
  bottom: the same with the latent alignment term. Bars average the tasks that
  have all eight conditions (AT and LAT for the active group); the gray line is
  the same encoder trained end-to-end; error bars as in
  Figure~\ref{fig:results-e2e-groups}.}
    \label{fig:results-pretrain-stacked}
\end{figure}
Odd--Even tests whether an encoding preserves the belief's parity-constrained, comb-shaped support rather than only its location and spread. Under end-to-end training, no encoder exceeds 57\% success, with PointNet performing best and the gradient CGF representation second. Reconstruction pretraining improves the frozen Set Transformer from 22\% to 62\%, but provides no consistent benefit across encoders or transfer modes. Task-specific pretraining on the categorical posterior produces the clearest improvement: all three frozen set encoders reach 87\% success, while their fine-tuned variants achieve 82--85\%. Unlike on the beacon tasks, the three set architectures perform similarly under this objective, suggesting that direct supervision helps each preserve the fine-grained support needed for exact prediction. The CGF representations gain little from pretraining, and latent alignment again provides no consistent improvement.

%% file: sections/conclusion.tex
\section{Conclusion}

We studied particle-belief encoding as a distinct representation-learning problem for RL, holding the particle filter and downstream learning pipeline fixed.
Three findings give concrete guidance.
First, pooling-based set encoders and gradient CGF features are the reliable choices when the encoder is trained end-to-end; the Set Transformer has the highest ceiling but reaches it only from a pretrained start.
Second, the pretraining objective matters more than the transfer mode: reconstruction helps selectively, whereas task-specific prediction is decisive when its target is the decision variable and can hurt when it replaces a belief that must stay uncertain with a point estimate. Once the objective matches the task, a frozen encoder performs about as well as a fine-tuned one.
Third, latent alignment gives no consistent benefit.
An encoding that reconstructs the belief well is therefore not necessarily a good input for control, and the encoder and its objective should be treated as part of the control design.
These conclusions rest on eight tasks and one RL algorithm under a shared training recipe.
Future work should develop task-agnostic objectives that preserve uncertainty and relational belief structure without privileged task-specific labels, including objectives trained jointly with RL, and test higher-dimensional partially observable problems and other RL algorithms.

%% file: sections/mandatory_ICLR_sections.tex
\newpage

\subsection*{AI use statement}

This statement follows the ICLR 2027 AI policy for authors, under which this work was submitted for review. We used generative AI tools to implement methods, including the code for the
POMDP environments that generate our training and evaluation data. The
environment specifications (state and observation spaces, dynamics, and reward
functions) were designed by the authors, and two authors verified the
implementations against them. We did not use generative AI tools to generate
synthetic datasets, develop theory, formulate or prove mathematical claims,
propose hypotheses, design experiments, or interpret results. We also used
generative AI tools to create and edit software code, edit the paper for
readability, identify relevant literature, and format references. We have
reviewed all AI-assisted work: two authors reviewed and tested all AI-assisted
code, every AI-suggested reference was checked against the original source,
and all AI-edited text was checked to ensure it reflects our methods and
claims. We take responsibility for the final content of this work.

\subsection*{Ethics statement}

This work does not involve human subjects. All datasets used are publicly
available and were used in accordance with their licenses. We are not aware of any conflicts of interest involving this work or its authors.

\subsection*{Reproducibility statement}
Sections~\ref{sec:methods} and~\ref{sec:experiments} describe the method and the experimental setup.
Appendices~\ref{app:anttag-details}, \ref{app:beacon-details}, and~\ref{app:oddeven-details} give the task parameters, and
Appendices~\ref{app:encoders}, \ref{app:ppo}, and~\ref{app:pretraining} give the encoder architectures, the PPO hyperparameters, and the pretraining details.
Experiments were run across several compute nodes with differing hardware
configurations. All experiments can be reproduced on one NVIDIA RTX 3080. We will publicly release the code and the trained model weights used
for the final results upon publication.

\subsection*{Acknowledgments}

We thank Jackson Wagner for meaningful discussions. This work was supported by NSF CAREER Award 2340958 and NSF NRI Award 2133141.

%% file: sections/appendix.tex
\newpage
\raggedbottom
\appendix
\section{Per-task results}
\label{app:per-task-results}

Tables~\ref{tab:e2e}--\ref{tab:task_aligned} give every cell behind
Figures~\ref{fig:results-e2e-groups} and~\ref{fig:results-pretrain-stacked}: one
row per task, one column per encoder and transfer mode, mean $\pm$ standard error
over PPO seeds. Cells with fewer than five seeds are marked and listed in the
captions. Beacon Sweep is reported here as the fraction of the five beacons
collected; the summaries in the main text use the rate of collecting all five.

\input{tables/results/table_e2e}
\input{tables/results/table_recon}
\input{tables/results/table_task}
\input{tables/results/table_recon_aligned}
\input{tables/results/table_task_aligned}

\section{Ant Tag variants: environment details}
\label{app:anttag-details}
\input{tables/results/table_anttag_params}

Table~\ref{tab:anttag-params} lists the three Ant Tag variants. All three
control the MuJoCo ant with eight torque actions in $[-1,1]$; the target is a
point that moves once per step by the listed rule. The particle filter runs the
same rule on 100 particles, is re-initialized every episode, and is the only
channel through which the target reaches the policy. Quantities that the
filter needs but the policy should not see, such as the parameters of the prior
used to initialize the particles, are passed to the filter and withheld from the
policy; the one exception is the frame-stacking baseline on Hideout Ant Tag,
which receives the den prior (Appendix~\ref{app:framestack}). During training the
visibility radius starts at 100, so the target is always visible while the ant
learns to walk, and shrinks to its evaluation value; the target's evasion
switches on over the same window, and the shaping terms follow the listed
schedules. Evaluation uses the environment's own reward with the end value of
every curriculum, over 300 episodes with deterministic actions. Hideout Ant Tag
is built so that the belief mean carries no information: the two active dens
lie on opposite sides of the origin with masses in inverse proportion to their
distances, so the policy has to read the posterior's shape to decide which den
to search, and entering either inactive candidate ends the episode at $-300$.

\section{Beacon tasks: environment details}
\label{app:beacon-details}
\input{tables/results/table_beacon_params}

Table~\ref{tab:beacon-params} lists the three beacon tasks. Survivor locations
$\mu_j$ are drawn uniformly in $[2,18]^2$ by rejection until every pair is at
least the minimum separation apart; the agent starts uniformly in the arena and
moves by $p_{t+1}=\mathrm{clip}(p_t+a_t)$, $a_t\in[-1,1]^2$. The belief handed
to the encoders is redrawn at every step: $n_j$ particles from
$\mathcal{N}(\mu_j,\sigma_j^2 I)$ per active beacon, clipped to the arena,
shuffled, equally weighted, and expressed relative to the agent (divided by the
arena half-width, 10). It is the stationary belief of a listener that clusters
and counts anonymous pings, each group kept at its fix spread; it is not a
posterior over $\mu_j$, which would sharpen with further pings. Component
identity, mass and spread, the decision variables of the three tasks, are exact
in it up to sampling noise, and no filtering error enters. Reward shaping is
potential-based (discount $0.99$) on the true centers, which the policy never
sees, and leaves the optimal policy unchanged; evaluation scores outcomes only,
over 300 episodes with five active beacons and contact radius $0.6$ (episode $i$
seeded $42+i$, deterministic actions). The curricula on the number of active
beacons and, for Beacon Sweep, on the contact radius are linear over the first
$40\%$ of training.

\section{Odd--Even: environment details}
\label{app:oddeven-details}
\input{tables/results/table_oddeven_params}

Table~\ref{tab:oddeven-params} lists the Odd--Even task. The belief is the exact
posterior, one weight per state updated by each observation's likelihood, so the
encoders see a categorical distribution with one particle per state and no
filtering error enters. The first observation already rules out every state of
the wrong parity, so the posterior is a comb from the first step on; what
changes over the episode is how sharply its mass concentrates within the parity,
and by step 21 the mode holds more than $0.9$ of it. We therefore score the
steady state, steps 22--30, where the optimal action is the posterior mode and
rounding the posterior mean names a zero-mass state in about a fifth of the
steps; evaluation uses 150 episodes (episode $i$ seeded $42+i$, deterministic
actions).

\section{Frame-stacking baseline}
\label{app:framestack}

The baseline has no encoder: the policy and value MLPs read the last $k$ raw
observation frames, concatenated newest-last and zero-padded at the start of an
episode (reset-frame padding on the Ant Tag variants), under each task's PPO
recipe, horizon and curriculum. The particle belief is built but not shown to
the policy. Table~\ref{tab:framestack-input} gives the frame of each family;
Table~\ref{tab:framestack-results} its results.
\input{tables/results/table_frame_stacking}
\input{tables/results/table_frame_stacking_family}

On the beacon tasks the raw observation is generated from the same hidden
layout as the belief: after the agent moves, beacon $j$ pings with probability
$\rho_j$ (Table~\ref{tab:beacon-params}), reporting $\mu_j+\sigma_j\epsilon$,
$\epsilon\sim\mathcal{N}(0,I)$, clipped to the arena; a reached beacon no
longer pings. The ping generator uses its own random stream, so layouts and
evaluation episodes are identical to the belief-encoder runs. Two facts qualify
this baseline's results. First, the belief encoders receive the converged
mixture from the first step, whereas frame stacking starts each episode with an
empty history, so the comparison favors the encoders; $k=60$ spans the whole
episode, and $k=20$ (and $k=1$ on Beacon Sweep) gave the same results. Second,
the input suffices: a hand-written rule that reads the newest frame and moves
toward the nearest ping reaches all five beacons in every one of the 300
evaluation episodes of Beacon Sweep, and scores $0.08$ on Faint Beacon and
$0.19$ on Diffuse Beacon because it ignores mass and spread. What the flat
policy fails to learn under PPO is a selection over slots whose order changes at
every step, which a set encoder provides by construction.

\section{Encoder architectures and sizes}
\label{app:encoders}

Table~\ref{tab:encoder-sizes} lists the output width and the number of trainable
parameters of every belief representation. Particle coordinates are divided by a
per-task length scale before they are encoded.

\noindent\textbf{Learned set networks:}
Each set network outputs eight codes of eight numbers, a 64-dimensional
representation. Deep Sets and PointNet apply a four-layer MLP of width 128 with
ReLU activations to every augmented particle $[x_i, Nw_i]$, pool the result, and
apply a second four-layer MLP of width 128. The Set Transformer uses two induced
set attention blocks with 16 inducing points, width 64, and four attention
heads, followed by attention pooling with eight seed vectors, two self-attention
blocks, and a linear layer that maps each pooled vector to eight numbers. The
widths are chosen so that the three set networks have nearly the same number of
trainable parameters (Table~\ref{tab:encoder-sizes}).

\noindent\textbf{CGF features:}
We use $m=64$ probes. For scalar particles, the probes are 32 magnitudes spaced
logarithmically between $0.25$ and a task-dependent maximum, each taken with
both signs. For planar particles, they are eight directions at multiples of
$45^\circ$ combined with eight magnitudes on the same logarithmic grid. Learned
probes start from these values. The features are passed to the policy directly,
without a readout layer.

\noindent\textbf{Statistical summaries:}
The Gaussian summary is the weighted mean, the per-coordinate variances, and the
covariance. The moment summary is the weighted mean and the central moments of
orders 2 to 8 of each coordinate.

\begin{table}[H]
\caption{Size of each belief representation: the width of the vector passed to the policy and the number of trainable parameters, for tasks with scalar particles (1-D: Car--Flag, Odd--Even) and with planar particles (2-D: all others). Every policy also receives the task's non-belief observation. A learned probe is one number in 1-D and three in 2-D (a magnitude and a direction vector). Frame stacking has no encoder; its width is that of the stacked observations. On the beacon tasks the actor and the critic hold separate copies of the encoder.}
\label{tab:encoder-sizes}
\begin{center}
\footnotesize
\setlength{\tabcolsep}{4pt}
\begin{tabular}{lllrr}
\toprule
 & \multicolumn{2}{c}{Output width} & \multicolumn{2}{c}{Trainable parameters} \\
\cmidrule(lr){2-3} \cmidrule(lr){4-5}
Representation & 1-D tasks & 2-D tasks & 1-D tasks & 2-D tasks \\
\midrule
Gaussian & 2 & 5 & 0 & 0 \\
Moments & 8 & 16 & 0 & 0 \\
\addlinespace[2pt]
$K$, fixed probes & 64 & 64 & 0 & 0 \\
$K$, learned probes & 64 & 64 & 64 & 192 \\
$\nabla K$, fixed probes & 64 & 128 & 0 & 0 \\
$\nabla K$, learned probes & 64 & 128 & 64 & 192 \\
\addlinespace[2pt]
Deep Sets & 64 & 64 & 107\,712 & 107\,840 \\
PointNet & 64 & 64 & 107\,712 & 107\,840 \\
Set Transformer & 64 & 64 & 109\,448 & 109\,640 \\
\addlinespace[2pt]
Frame stacking & 480 (CF), 60 (OE) & 248--254 (Ant Tag), 1\,020 (beacons) & 0 & 0 \\
\bottomrule
\end{tabular}
\end{center}
\end{table}

\section{Reinforcement learning hyperparameters}
\label{app:ppo}

All policies are trained with the PPO implementation of
Stable-Baselines3~\citep{raffin2021sb3}. Table~\ref{tab:ppo} lists the settings,
which are the same for every encoder and training regime within a task. The
policy and value networks receive the non-belief observation concatenated with
the belief representation. A frozen encoder receives no gradient. A fine-tuned
encoder is placed in its own parameter group at $0.1$ times the policy learning
rate and follows the same schedule.

\begin{table}[H]
\caption{PPO hyperparameters. Normalization refers to running normalization of the non-belief observation and of the training reward; particles and weights are never normalized this way.}
\label{tab:ppo}
\begin{center}
\footnotesize
\setlength{\tabcolsep}{5pt}
\begin{tabular}{llll}
\toprule
Hyperparameter & Car--Flag, Odd--Even & Ant Tag variants & Beacon tasks \\
\midrule
Training steps & 1M (CF), 3M (OE) & 6M & 3M \\
Learning rate & $3\times10^{-4}$, linear decay & $3\times10^{-4}$, linear decay & $3\times10^{-4}$, constant \\
Parallel environments & 4 & 4 & 16 \\
Steps per environment per update & 2048 & 4096 & 256 \\
Minibatch size & 64 & 64 & 1024 \\
Epochs per update & 10 & 10 & 10 \\
Discount $\gamma$ & 0.99 & 0.99 & 0.99 \\
GAE parameter & 0.95 & 0.95 & 0.95 \\
Clip range & 0.2 & 0.2 & 0.2 \\
Target KL divergence & 0.03 & 0.03 & 0.05 \\
Entropy coefficient & 0 & 0 & 0.005 \\
Value-loss coefficient & 0.5 & 0.5 & 0.5 \\
Maximum gradient norm & 0.5 & 0.5 & 0.5 \\
Policy and value networks & $2\times64$, tanh & $2\times64$, tanh & $2\times128$, tanh \\
Encoder for actor and critic & shared & shared & separate copies \\
Normalization & yes & yes & no \\
\bottomrule
\end{tabular}
\end{center}
\end{table}

\section{Pretraining details}
\label{app:pretraining}

\noindent\textbf{Data:}
Table~\ref{tab:pretrain-data} lists the source of the pretraining beliefs for
each task. Each stored belief is a full weighted particle set produced by the
task's own belief source. On the Ant Tag variants, each episode uses one of four
behaviors: walking between sampled waypoints (probability $0.5$), pursuing the
belief mean ($0.3$), pursuing the target with the target always visible ($0.1$),
or taking random actions ($0.1$); the first three drive a separately trained
locomotion policy. On Beacon Sweep, $60\%$ of the episodes pursue an active
beacon and $40\%$ follow a random walk. For Faint and Diffuse Beacon,
reconstruction pretraining draws new beliefs from the task's reset distribution
at every epoch instead of using rollouts. Task-specific pretraining on Diffuse
Beacon combines beliefs visited by previously trained policies with freshly
drawn ones. In Odd--Even the belief does not depend on the actions, so random
actions suffice. Car--Flag is not pretrained.

\begin{table}[H]
\caption{Source and number of pretraining beliefs. ``Reset distribution'' means that beliefs are drawn directly from the task's initial-layout distribution, without rollouts.}
\label{tab:pretrain-data}
\begin{center}
\footnotesize
\setlength{\tabcolsep}{5pt}
\begin{tabular}{lp{5.4cm}p{4.6cm}}
\toprule
Task & Source & Beliefs \\
\midrule
Ant Tag & scripted rollouts & 100\,000 \\
Lantern Ant Tag & scripted rollouts & 100\,000 \\
Hideout Ant Tag & scripted rollouts & 38\,271 \\
Beacon Sweep & scripted rollouts & 185\,009 \\
Faint Beacon & reset distribution & 100\,000 per epoch (reconstruction); 300\,000 per epoch (task-specific) \\
Diffuse Beacon & reset distribution (reconstruction); rollouts of trained policies and reset distribution (task-specific) & 100\,000 per epoch (reconstruction); 161\,960 and 87\,092 per epoch (task-specific) \\
Odd--Even & rollouts with random actions & 136\,400 \\
\bottomrule
\end{tabular}
\end{center}
\end{table}

\noindent\textbf{Reconstruction:}
The decoder outputs $N$ particle locations with uniform weights $1/N$. The loss
is the debiased Sinkhorn divergence between this uniformly weighted set and the
input particles carrying their filter weights. The input weights therefore enter
as the target marginal of the transport problem, and the input belief is not
resampled. A weighted belief is thus reconstructed by an equally weighted one
that places more particles where the input has more mass, as resampling does.
We use a squared Euclidean cost with blur $0.01$ on the Ant Tag variants and
$0.02$ elsewhere, in coordinates divided by the task's length scale. Training
uses Adam with a cosine learning-rate schedule and gradient-norm clipping at
$1.0$, and keeps the checkpoint with the lowest validation loss.

\noindent\textbf{Task-specific prediction:}
The position and offset labels of Section~\ref{sec:experiments:comparisons} are read
from simulator state that is hidden from the policy. The posterior labels in
Odd--Even and Hideout Ant Tag are computed from the belief itself.
Reconstruction needs no labels.
The prediction head is a three-layer MLP of width 256 (a single linear layer in
Odd--Even), trained with mean-squared error for position targets and
cross-entropy for categorical targets. Training uses Adam with gradient-norm
clipping at $1.0$. The learning rate is halved when the validation loss
plateaus, and training stops after 15 epochs without improvement; Odd--Even
instead uses a cosine schedule for a fixed number of epochs. The checkpoint with
the lowest validation loss is kept.

\noindent\textbf{Latent alignment:}
The alignment term of Section~\ref{sec:methods:pretraining} is computed over
2\,016 sampled pairs of beliefs per batch. Its weight is zero for the first two
epochs and then increases linearly to $\lambda$ over the next three; for the set
networks under task-specific pretraining on Ant Tag and Lantern Ant Tag it is
applied at full weight from the first epoch. The term is used only during
pretraining and does not enter checkpoint selection.
Table~\ref{tab:pretrain-settings} lists $\lambda$ and the optimization settings.

\begin{table}[H]
\caption{Pretraining settings. ``Up to'' marks objectives with early stopping. $\lambda$ is the alignment weight used when the alignment term is added. $^{\ast}$On Ant Tag and Lantern Ant Tag, reconstruction without alignment uses batch size 32 and 100 epochs for the CGF encoders.}
\label{tab:pretrain-settings}
\begin{center}
\footnotesize
\setlength{\tabcolsep}{5pt}
\begin{tabular}{llllll}
\toprule
Task & Objective & Batch size & Epochs & Learning rate & $\lambda$ \\
\midrule
Ant Tag, Lantern Ant Tag & reconstruction & 128$^{\ast}$ & 200$^{\ast}$ & $10^{-3}$ & 0.2 \\
 & task-specific & 256 & up to 120 & $10^{-3}$ & 0.2 \\
Hideout Ant Tag & reconstruction & 32 & 100 & $10^{-3}$ & -- \\
 & task-specific & 256 & up to 120 & $10^{-3}$ & -- \\
Beacon Sweep & reconstruction & 256 & 200 & $2.8\times10^{-3}$ & 0.02 \\
 & task-specific & 256 & up to 120 & $10^{-3}$ & 0.02 \\
Faint Beacon & reconstruction & 256 & 200 & $2.8\times10^{-3}$ & 0.02 \\
 & task-specific & 256 & up to 60 & $10^{-3}$ & 0.02 \\
Diffuse Beacon & reconstruction & 256 & 200 & $2.8\times10^{-3}$ & 0.02 \\
 & task-specific & 256 & up to 100 & $10^{-3}$ & 0.02 \\
Odd--Even & reconstruction & 32 & 100 & $10^{-3}$ & 0.02 \\
 & task-specific & 512 & 200 & $10^{-3}$ & 0.2 \\
\bottomrule
\end{tabular}
\end{center}
\end{table}

\section{Ablation}
\label{app:ablation}

Figures~\ref{fig:ablation-moment-order} and~\ref{fig:ablation-cgf-probes}
vary the highest moment order of the moment encoder and the number of probes of
the CGF encoders on Faint Beacon and Diffuse Beacon, under end-to-end training
with all other settings unchanged. Performance does not improve systematically
with either quantity. Every setting stays near the $20\%$ chance rate; the
highest value is $26\%$, for gradient CGF features with 64 learned probes on
Diffuse Beacon, which is the setting used in the main text.

\input{figures/ablate}

%% file: tables/results/table_e2e.tex
\providecommand{\ms}[2]{#1\,{\fontsize{5}{6}\selectfont$\pm$#2}}
\providecommand{\best}[1]{\textbf{#1}}
\providecommand{\second}[1]{\underline{#1}}

\begin{table}[H]
\caption{End-to-end training: each encoder is trained from a random initialization by PPO alone. Mean $\pm$ standard error over five PPO seeds, each seed evaluated on 300 episodes. Best per row in bold, second best underlined. Car--Flag: rate of episodes ending at the rewarding end; Ant Tag variants: rate of episodes in which the target is tagged; Odd--Even: exact-match rate over the steady-state steps 22--30; Beacon Sweep: fraction of the five beacons collected; Faint Beacon and Diffuse Beacon: rate of episodes that reach the correct beacon. Frame stack: the last $k$ raw observation frames ($k=160$ on Car--Flag, 8 on the Ant Tag variants, 30 on Odd--Even, 60 on the beacon tasks). Moments: central moments of orders 1--8.}
\label{tab:e2e}
\begin{center}
\scriptsize
\setlength{\tabcolsep}{1.8pt}
\begin{tabular}{lcccccccccc}
\toprule
 & \multicolumn{2}{c}{Statistics} & \multicolumn{4}{c}{CGF} & \multicolumn{3}{c}{Set networks} & \multicolumn{1}{c}{History} \\
\cmidrule(lr){2-3} \cmidrule(lr){4-7} \cmidrule(lr){8-10} \cmidrule(lr){11-11}
Task & Gaussian & Moments & $K$ fixed & $K$ learned & $\nabla K$ fixed & $\nabla K$ learned & Deep Sets & PointNet & Set Transf. & Frame Stacking \\
\midrule
Car--Flag & \ms{0.80}{0.12} & \ms{0.70}{0.12} & \ms{0.70}{0.12} & \ms{0.70}{0.12} & \ms{\second{0.90}}{0.10} & \ms{\second{0.90}}{0.10} & \ms{\best{1.00}}{0.00} & \ms{\best{1.00}}{0.00} & \ms{\best{1.00}}{0.00} & \ms{0.80}{0.12} \\
Ant Tag & \ms{0.55}{0.06} & \ms{0.18}{0.05} & \ms{0.48}{0.02} & \ms{0.58}{0.07} & \ms{0.74}{0.02} & \ms{0.75}{0.01} & \ms{\second{0.85}}{0.05} & \ms{0.69}{0.05} & \ms{\best{0.87}}{0.02} & \ms{0.02}{0.01} \\
Lantern Ant Tag & \ms{0.32}{0.03} & \ms{0.04}{0.01} & \ms{0.09}{0.03} & \ms{0.17}{0.02} & \ms{0.38}{0.04} & \ms{0.36}{0.03} & \ms{\best{0.59}}{0.04} & \ms{0.29}{0.02} & \ms{\second{0.43}}{0.05} & \ms{0.00}{0.00} \\
Hideout Ant Tag & \ms{0.29}{0.05} & \ms{0.72}{0.01} & \ms{0.71}{0.01} & \ms{0.72}{0.01} & \ms{0.70}{0.04} & \ms{\second{0.73}}{0.01} & \ms{\best{0.75}}{0.01} & \ms{\best{0.75}}{0.00} & \ms{\second{0.73}}{0.02} & \ms{\second{0.73}}{0.01} \\
Odd--Even & \ms{0.31}{0.02} & \ms{0.29}{0.02} & \ms{0.30}{0.01} & \ms{0.31}{0.01} & \ms{\second{0.42}}{0.03} & \ms{\second{0.42}}{0.02} & \ms{0.31}{0.03} & \ms{\best{0.57}}{0.01} & \ms{0.22}{0.04} & \ms{0.26}{0.01} \\
Beacon Sweep & \ms{0.22}{0.03} & \ms{0.19}{0.04} & \ms{0.86}{0.01} & \ms{0.89}{0.00} & \ms{0.92}{0.01} & \ms{0.94}{0.01} & \ms{\second{0.97}}{0.01} & \ms{\best{0.98}}{0.01} & \ms{0.21}{0.01} & \ms{0.24}{0.00} \\
Faint Beacon & \ms{0.13}{0.01} & \ms{0.12}{0.01} & \ms{\best{0.21}}{0.01} & \ms{0.19}{0.01} & \ms{0.19}{0.01} & \ms{\second{0.20}}{0.01} & \ms{0.18}{0.01} & \ms{0.15}{0.01} & \ms{0.12}{0.01} & \ms{0.12}{0.01} \\
Diffuse Beacon & \ms{0.13}{0.01} & \ms{0.13}{0.01} & \ms{0.19}{0.01} & \ms{0.17}{0.01} & \ms{\second{0.27}}{0.01} & \ms{0.26}{0.01} & \ms{0.15}{0.01} & \ms{\best{0.30}}{0.03} & \ms{0.11}{0.01} & \ms{0.14}{0.01} \\
\bottomrule
\end{tabular}
\end{center}
\end{table}

%% file: tables/results/table_recon.tex
\providecommand{\ms}[2]{#1\,{\fontsize{5}{6}\selectfont$\pm$#2}}
\providecommand{\best}[1]{\textbf{#1}}
\providecommand{\second}[1]{\underline{#1}}

\begin{table}[H]
\caption{Reconstruction: reconstruction pretraining (weighted Sinkhorn autoencoder). One pretraining run (seed 0) initializes every PPO seed; the encoder is then either frozen (left) or fine-tuned at $0.1\times$ the PPO learning rate (right). Mean $\pm$ standard error over five PPO seeds, each seed evaluated on 300 episodes. Best per row in bold, second best underlined. Metrics as in Table~\ref{tab:e2e}. Not run: Car--Flag (Car--Flag has only three reachable beliefs, nothing to pretrain on).}
\label{tab:recon}
\begin{center}
\scriptsize
\setlength{\tabcolsep}{2pt}
\begin{tabular}{lcccccccccc}
\toprule
 & \multicolumn{5}{c}{Frozen encoder} & \multicolumn{5}{c}{Fine-tuned encoder ($0.1\times$ LR)} \\
\cmidrule(lr){2-6} \cmidrule(lr){7-11}
Task & CGF $K$ & CGF $\nabla K$ & Deep Sets & PointNet & Set Transf. & CGF $K$ & CGF $\nabla K$ & Deep Sets & PointNet & Set Transf. \\
\midrule
Ant Tag & \ms{0.31}{0.05} & \ms{0.70}{0.02} & \ms{\best{0.94}}{0.03} & \ms{\second{0.90}}{0.02} & \ms{0.79}{0.03} & \ms{0.38}{0.03} & \ms{0.78}{0.03} & \ms{0.81}{0.05} & \ms{0.77}{0.05} & \ms{0.78}{0.03} \\
Lantern Ant Tag & \ms{0.09}{0.02} & \ms{0.40}{0.04} & \ms{0.38}{0.08} & \ms{0.35}{0.04} & \ms{0.40}{0.05} & \ms{0.09}{0.03} & \ms{0.36}{0.02} & \ms{\second{0.43}}{0.04} & \ms{0.34}{0.03} & \ms{\best{0.44}}{0.05} \\
Hideout Ant Tag & \ms{0.71}{0.02} & \ms{0.72}{0.02} & \ms{0.72}{0.01} & \ms{\best{0.74}}{0.01} & \ms{\second{0.73}}{0.02} & \ms{\best{0.74}}{0.01} & \ms{\second{0.73}}{0.01} & \ms{0.72}{0.02} & \ms{0.69}{0.06} & \ms{\best{0.74}}{0.01} \\
Odd--Even & \ms{0.29}{0.01} & \ms{0.44}{0.01} & \ms{0.48}{0.02} & \ms{0.48}{0.02} & \ms{\best{0.62}}{0.01} & \ms{0.28}{0.01} & \ms{0.43}{0.01} & \ms{0.54}{0.01} & \ms{\second{0.55}}{0.03} & \ms{0.40}{0.02} \\
Beacon Sweep & \ms{0.88}{0.01} & \ms{0.95}{0.01} & \ms{0.71}{0.01} & \ms{0.91}{0.00} & \ms{0.87}{0.03} & \ms{0.89}{0.02} & \ms{0.95}{0.01} & \ms{\second{0.97}}{0.01} & \ms{\best{0.99}}{0.00} & \ms{0.95}{0.01} \\
Faint Beacon & \ms{0.20}{0.02} & \ms{\best{0.24}}{0.01} & \ms{0.13}{0.01} & \ms{0.17}{0.01} & \ms{0.13}{0.02} & \ms{0.19}{0.01} & \ms{0.21}{0.01} & \ms{\second{0.23}}{0.01} & \ms{0.18}{0.01} & \ms{0.14}{0.01} \\
Diffuse Beacon & \ms{0.19}{0.00} & \ms{0.25}{0.02} & \ms{0.14}{0.01} & \ms{0.19}{0.02} & \ms{0.13}{0.01} & \ms{0.16}{0.01} & \ms{\best{0.27}}{0.02} & \ms{0.16}{0.01} & \ms{\second{0.26}}{0.01} & \ms{0.18}{0.01} \\
\bottomrule
\end{tabular}
\end{center}
\end{table}

%% file: tables/results/table_task.tex
\providecommand{\ms}[2]{#1\,{\fontsize{5}{6}\selectfont$\pm$#2}}
\providecommand{\best}[1]{\textbf{#1}}
\providecommand{\second}[1]{\underline{#1}}

\begin{table}[H]
\caption{Task-specific: task-specific pretraining (a supervised head on the task-relevant label). One pretraining run (seed 0) initializes every PPO seed; the encoder is then either frozen (left) or fine-tuned at $0.1\times$ the PPO learning rate (right). Mean $\pm$ standard error over five PPO seeds, each seed evaluated on 300 episodes. Best per row in bold, second best underlined. Metrics as in Table~\ref{tab:e2e}. Not run: Car--Flag (Car--Flag has only three reachable beliefs, nothing to pretrain on).}
\label{tab:task}
\begin{center}
\scriptsize
\setlength{\tabcolsep}{2pt}
\begin{tabular}{lcccccccccc}
\toprule
 & \multicolumn{5}{c}{Frozen encoder} & \multicolumn{5}{c}{Fine-tuned encoder ($0.1\times$ LR)} \\
\cmidrule(lr){2-6} \cmidrule(lr){7-11}
Task & CGF $K$ & CGF $\nabla K$ & Deep Sets & PointNet & Set Transf. & CGF $K$ & CGF $\nabla K$ & Deep Sets & PointNet & Set Transf. \\
\midrule
Ant Tag & \ms{0.43}{0.08} & \ms{0.71}{0.04} & \ms{0.86}{0.01} & \ms{0.78}{0.02} & \ms{0.87}{0.03} & \ms{0.47}{0.07} & \ms{0.68}{0.05} & \ms{\second{0.88}}{0.03} & \ms{0.80}{0.00} & \ms{\best{0.89}}{0.02} \\
Lantern Ant Tag & \ms{0.10}{0.02} & \ms{0.33}{0.03} & \ms{0.30}{0.06} & \ms{0.31}{0.02} & \ms{0.36}{0.02} & \ms{0.08}{0.01} & \ms{\second{0.39}}{0.02} & \ms{0.26}{0.05} & \ms{0.31}{0.03} & \ms{\best{0.45}}{0.06} \\
Hideout Ant Tag & \ms{0.61}{0.08} & \ms{0.74}{0.00} & \ms{0.73}{0.01} & \ms{0.72}{0.02} & \ms{0.71}{0.01} & \ms{0.73}{0.01} & \ms{0.74}{0.01} & \ms{\second{0.76}}{0.01} & \ms{\best{0.77}}{0.02} & \ms{0.71}{0.02} \\
Odd--Even & \ms{0.24}{0.02} & \ms{0.43}{0.01} & \ms{\best{0.87}}{0.00} & \ms{\best{0.87}}{0.00} & \ms{\best{0.87}}{0.00} & \ms{0.25}{0.02} & \ms{0.44}{0.01} & \ms{0.82}{0.01} & \ms{\second{0.85}}{0.01} & \ms{0.83}{0.02} \\
Beacon Sweep & \ms{0.86}{0.01} & \ms{0.91}{0.01} & \ms{\second{0.99}}{0.00} & \ms{\best{1.00}}{0.00} & \ms{\best{1.00}}{0.00} & \ms{0.87}{0.01} & \ms{0.92}{0.00} & \ms{\best{1.00}}{0.00} & \ms{\best{1.00}}{0.00} & \ms{\best{1.00}}{0.00} \\
Faint Beacon & \ms{0.18}{0.01} & \ms{0.22}{0.01} & \ms{0.40}{0.01} & \ms{0.25}{0.01} & \ms{\best{0.88}}{0.01} & \ms{0.20}{0.01} & \ms{0.22}{0.01} & \ms{0.40}{0.01} & \ms{0.34}{0.00} & \ms{\second{0.87}}{0.01} \\
Diffuse Beacon & \ms{0.19}{0.02} & \ms{0.21}{0.01} & \ms{0.63}{0.01} & \ms{0.79}{0.01} & \ms{\second{0.91}}{0.00} & \ms{0.20}{0.01} & \ms{0.23}{0.01} & \ms{0.52}{0.04} & \ms{0.78}{0.01} & \ms{\best{0.92}}{0.01} \\
\bottomrule
\end{tabular}
\end{center}
\end{table}

%% file: tables/results/table_recon_aligned.tex
\providecommand{\ms}[2]{#1\,{\fontsize{5}{6}\selectfont$\pm$#2}}
\providecommand{\best}[1]{\textbf{#1}}
\providecommand{\second}[1]{\underline{#1}}

\begin{table}[H]
\caption{Reconstruction + alignment: reconstruction pretraining with the latent metric alignment term. One pretraining run (seed 0) initializes every PPO seed; the encoder is then either frozen (left) or fine-tuned at $0.1\times$ the PPO learning rate (right). Mean $\pm$ standard error over five PPO seeds, each seed evaluated on 300 episodes. Best per row in bold, second best underlined. Metrics as in Table~\ref{tab:e2e}. Not run: Car--Flag, Hideout Ant Tag (Car--Flag has only three reachable beliefs, nothing to pretrain on; Hideout Ant Tag has no alignment runs).}
\label{tab:recon_aligned}
\begin{center}
\scriptsize
\setlength{\tabcolsep}{2pt}
\begin{tabular}{lcccccccccc}
\toprule
 & \multicolumn{5}{c}{Frozen encoder} & \multicolumn{5}{c}{Fine-tuned encoder ($0.1\times$ LR)} \\
\cmidrule(lr){2-6} \cmidrule(lr){7-11}
Task & CGF $K$ & CGF $\nabla K$ & Deep Sets & PointNet & Set Transf. & CGF $K$ & CGF $\nabla K$ & Deep Sets & PointNet & Set Transf. \\
\midrule
Ant Tag & \ms{0.66}{0.05} & \ms{0.75}{0.03} & \ms{0.79}{0.08} & \ms{\second{0.87}}{0.04} & \ms{0.86}{0.05} & \ms{0.72}{0.04} & \ms{0.74}{0.05} & \ms{\best{0.93}}{0.01} & \ms{0.83}{0.02} & \ms{0.76}{0.05} \\
Lantern Ant Tag & \ms{0.40}{0.03} & \ms{0.36}{0.03} & \ms{\second{0.50}}{0.08} & \ms{0.33}{0.01} & \ms{0.43}{0.06} & \ms{0.38}{0.04} & \ms{0.39}{0.02} & \ms{0.41}{0.05} & \ms{0.42}{0.02} & \ms{\best{0.53}}{0.06} \\
Odd--Even & \ms{0.30}{0.03} & \ms{0.35}{0.02} & \ms{\second{0.55}}{0.02} & \ms{0.47}{0.03} & \ms{0.51}{0.01} & \ms{0.31}{0.02} & \ms{0.39}{0.02} & \ms{0.48}{0.05} & \ms{\best{0.56}}{0.02} & \ms{0.50}{0.04} \\
Beacon Sweep & \ms{0.88}{0.01} & \ms{0.96}{0.00} & \ms{0.66}{0.01} & \ms{0.88}{0.00} & \ms{0.82}{0.06} & \ms{0.89}{0.00} & \ms{0.96}{0.01} & \ms{\second{0.97}}{0.01} & \ms{\best{1.00}}{0.00} & \ms{\second{0.97}}{0.01} \\
Faint Beacon & \ms{0.16}{0.01} & \ms{\second{0.22}}{0.01} & \ms{0.12}{0.01} & \ms{0.17}{0.01} & \ms{0.16}{0.01} & \ms{0.19}{0.01} & \ms{\second{0.22}}{0.01} & \ms{\best{0.23}}{0.01} & \ms{0.16}{0.00} & \ms{\second{0.22}}{0.02} \\
Diffuse Beacon & \ms{0.19}{0.02} & \ms{\second{0.23}}{0.02} & \ms{0.13}{0.01} & \ms{0.17}{0.00} & \ms{0.19}{0.01} & \ms{0.19}{0.01} & \ms{\second{0.23}}{0.01} & \ms{0.15}{0.01} & \ms{\best{0.26}}{0.02} & \ms{0.22}{0.01} \\
\bottomrule
\end{tabular}
\end{center}
\end{table}

%% file: tables/results/table_task_aligned.tex
\providecommand{\ms}[2]{#1\,{\fontsize{5}{6}\selectfont$\pm$#2}}
\providecommand{\best}[1]{\textbf{#1}}
\providecommand{\second}[1]{\underline{#1}}

\begin{table}[H]
\caption{Task-specific + alignment: task-specific pretraining with the latent metric alignment term. One pretraining run (seed 0) initializes every PPO seed; the encoder is then either frozen (left) or fine-tuned at $0.1\times$ the PPO learning rate (right). Mean $\pm$ standard error over five PPO seeds, each seed evaluated on 300 episodes. Best per row in bold, second best underlined. Metrics as in Table~\ref{tab:e2e}. Not run: Car--Flag, Hideout Ant Tag (Car--Flag has only three reachable beliefs, nothing to pretrain on; Hideout Ant Tag has no alignment runs).}
\label{tab:task_aligned}
\begin{center}
\scriptsize
\setlength{\tabcolsep}{2pt}
\begin{tabular}{lcccccccccc}
\toprule
 & \multicolumn{5}{c}{Frozen encoder} & \multicolumn{5}{c}{Fine-tuned encoder ($0.1\times$ LR)} \\
\cmidrule(lr){2-6} \cmidrule(lr){7-11}
Task & CGF $K$ & CGF $\nabla K$ & Deep Sets & PointNet & Set Transf. & CGF $K$ & CGF $\nabla K$ & Deep Sets & PointNet & Set Transf. \\
\midrule
Ant Tag & \ms{0.48}{0.02} & \ms{0.68}{0.04} & \ms{0.79}{0.04} & \ms{0.77}{0.05} & \ms{0.81}{0.03} & \ms{0.48}{0.06} & \ms{0.79}{0.03} & \ms{0.87}{0.01} & \ms{\second{0.89}}{0.02} & \ms{\best{0.91}}{0.02} \\
Lantern Ant Tag & \ms{0.11}{0.01} & \ms{0.34}{0.04} & \ms{\second{0.41}}{0.03} & \ms{0.32}{0.02} & \ms{0.37}{0.00} & \ms{0.10}{0.01} & \ms{0.33}{0.04} & \ms{\best{0.49}}{0.04} & \ms{0.32}{0.01} & \ms{0.34}{0.04} \\
Odd--Even & \ms{0.28}{0.01} & \ms{0.44}{0.02} & \ms{\second{0.87}}{0.00} & \ms{0.86}{0.01} & \ms{\best{0.88}}{0.00} & \ms{0.28}{0.02} & \ms{0.45}{0.01} & \ms{0.81}{0.01} & \ms{0.79}{0.01} & \ms{0.86}{0.01} \\
Beacon Sweep & \ms{0.87}{0.01} & \ms{0.94}{0.01} & \ms{\second{0.99}}{0.00} & \ms{\best{1.00}}{0.00} & \ms{\second{0.99}}{0.01} & \ms{0.86}{0.02} & \ms{0.95}{0.01} & \ms{\best{1.00}}{0.00} & \ms{\best{1.00}}{0.00} & \ms{\best{1.00}}{0.00} \\
Faint Beacon & \ms{0.17}{0.01} & \ms{0.20}{0.01} & \ms{0.31}{0.01} & \ms{0.25}{0.00} & \ms{\second{0.88}}{0.01} & \ms{0.20}{0.02} & \ms{0.18}{0.01} & \ms{0.32}{0.01} & \ms{0.31}{0.01} & \ms{\best{0.89}}{0.01} \\
Diffuse Beacon & \ms{0.22}{0.01} & \ms{0.19}{0.01} & \ms{0.64}{0.06} & \ms{0.74}{0.01} & \ms{\second{0.90}}{0.01} & \ms{0.22}{0.01} & \ms{0.22}{0.01} & \ms{0.51}{0.02} & \ms{0.74}{0.01} & \ms{\best{0.92}}{0.01} \\
\bottomrule
\end{tabular}
\end{center}
\end{table}

%% file: tables/results/table_anttag_params.tex
\begin{table}[H]
\caption{Ant Tag task parameters. $d$ is the ant--target distance,
$r_{\mathrm{vis}}$ and $r_{\mathrm{tag}}$ the visibility and tag radii,
$u=\mathrm{clip}\big((r_{\mathrm{vis}}-d)/(r_{\mathrm{vis}}-r_{\mathrm{tag}}),0,1\big)\,e$
the target's urgency with evasion scale $e$, and $H(w)$ the entropy of the
particle weights. Curricula are linear over the stated fraction of training;
evaluation uses the end values. Lengths are in arena units (the ant's torso has
radius $0.25$).}
\label{tab:anttag-params}
\begin{center}
\small
\setlength{\tabcolsep}{4pt}
\begin{tabular}{p{2.4cm}p{3.0cm}p{3.0cm}p{4.1cm}}
\toprule
 & Ant Tag & Lantern Ant Tag & Hideout Ant Tag \\
\midrule
Episode length & 400 & 400 & 300 \\
Cage half-width & 4.5 & 4.5 & 7.0 \\
$r_{\mathrm{vis}}$ (train $\to$ eval) & $100\to3.0$ (20--50\%) & $100\to1.5$ (20--40\%) & $100\to1.0$ (20--50\%) \\
$r_{\mathrm{tag}}$ & 1.5 & 1.0 & 0.6 \\
Target step & 0.5 & 0.3 & 0.5 \\
Evasion scale $e$ & \multicolumn{3}{p{106.6mm}}{$0\to1$ over 20--50\% of training; $1$ at evaluation} \\
Target motion & \multicolumn{2}{p{62.8mm}}{one move per step relative to the ant: away with probability $0.25+0.45u$, stay $0.25-0.20u$, left or right the rest; slides along the cage wall} & leashed to a disc of radius $0.4$ around its den: away $0.25+0.45u$, toward $0.25(1-u)$, stay $0.05u$, left or right the rest; released into the Ant Tag rule over the whole cage once the ant comes within $1.4$ of the empty active den \\
Dens & \multicolumn{2}{c}{--} & four candidate centers on the diagonal at distances $2.4$ (near) and $6.75$ (far); a mirror bit makes one near den and the opposite far den active; the target is in the near den with probability $6.75/(2.4+6.75)=0.74$, which pins the prior mean at the origin; the two inactive candidates are hazards \\
Spawn & \multicolumn{2}{p{62.8mm}}{ant and target uniform in the cage, at least $5.0$ apart} & ant uniform in a disc of radius $0.5$ at the center; target uniform in its den's disc \\
\raggedright Initial particles (100) & \multicolumn{2}{p{62.8mm}}{uniform over the cage beyond $5.0$ from the ant, equal weights} & mass $0.37$ on each near candidate and $0.13$ on each far one; resolved to the two active dens at the first update \\
Filter update & \multicolumn{3}{p{106.6mm}}{the target's own motion rule, no added noise; a sighting multiplies weights by $\mathcal{N}(z;\,x,\,0.1^2 I)$, no sighting multiplies particles inside $r_{\mathrm{vis}}$ by $0.1$; partial resampling when the effective sample size drops below $50$, with jitter of half the target step; Hideout Ant Tag also multiplies by $10^{-6}$ the particles that a silent or triggered spook rules out} \\
Reward & \multicolumn{2}{p{62.8mm}}{$-1$ per step, $0$ on the tag; no timeout penalty} & $-1$ per step, $0$ on the tag, $-300$ on entering a hazard \\
\raggedright Shaping (training only) & \multicolumn{2}{p{62.8mm}}{$-c_d\,\|\mathrm{ant}-\mathrm{belief\ mean}\|-2H(w)+$ tag bonus; $c_d$ $1\to0.15$ and bonus $0\to50$ over 20--50\%} & $-c_d\,\|\mathrm{ant}-\mathrm{belief\ mean}\|+$ tag bonus; $c_d$ $1\to0$ and bonus $0\to50$ over 30--70\% \\
Termination & tag, or timeout & tag, or timeout & tag, hazard, or timeout \\
\raggedright Policy observation & \multicolumn{3}{p{106.6mm}}{the 31-d MuJoCo state of the ant (positions and velocities) with the target coordinates zeroed; the target reaches the policy only through the belief} \\
Evaluation metric & tag rate & tag rate & tag rate; a hazard is a failure \\
\bottomrule
\end{tabular}
\end{center}
\end{table}

%% file: tables/results/table_beacon_params.tex
\begin{table}[H]
\caption{Beacon task parameters. Particle counts $n_j$ sum to 100 over the
active beacons; $\rho_j$ is the ping probability, seen only by the
frame-stacking baseline; $d(\cdot)$ is the agent's distance to a location.}
\label{tab:beacon-params}
\begin{center}
\small
\setlength{\tabcolsep}{4pt}
\begin{tabular}{lp{2.5cm}p{3.3cm}p{3.2cm}}
\toprule
 & Beacon Sweep & Faint Beacon & Diffuse Beacon \\
\midrule
Episode length & 60 & 60 & 60 \\
Min.\ separation of $\mu_j$ & 2.5 & 3.0 & 3.0 \\
Fix scatter $\sigma_j$ & $U[0.3,0.7]$ & $U[0.3,0.7]$ & $U[0.3,0.9]$; the widest exceeds the next by $\ge 0.2$ \\
Particle counts $n_j$ & equal over live beacons & unique minimum $n_{\min}\sim U\{6,\dots,15\}$; every other $\ge n_{\min}+5$ & equal \\
Ping probability $\rho_j$ & 1 for live beacons & $n_j/\max_k n_k$ & 1 \\
Target & every live beacon & $\arg\min_j n_j$ & $\arg\max_j \sigma_j$ \\
Active beacons (train $\to$ eval) & $1\to5$ & $2\to5$ & $2\to5$ \\
Contact radius (train $\to$ eval) & $1.6\to0.6$ & $0.6$ & $0.6$ \\
Termination & all reached, or timeout & first contact, or timeout & first contact, or timeout \\
Reward & $+10$ per beacon reached; $-0.01$ per step & $+20$ target, $-5$ other, $-40$ timeout; $-0.01$ per step & $+20$ target, $-5$ other, $-20$ timeout; $-0.01$ per step \\
Shaping potential & $20\,(\#\mathrm{reached}) - d(\mathrm{nearest\ live\ }\mu_j)$ & $-d(\mathrm{target\ }\mu_j)$ & $-d(\mathrm{target\ }\mu_j)$ \\
Evaluation metric & fraction reached; all-five rate & correct / wrong / timeout & correct / wrong / timeout \\
\bottomrule
\end{tabular}
\end{center}
\end{table}

%% file: tables/results/table_oddeven_params.tex
\begin{table}[H]
\caption{Odd--Even task parameters. $s^\ast$ is the hidden integer, $o_t$ the
observation drawn at step $t$ and $b_t$ the belief; the belief is the exact
posterior, so no filtering error enters.}
\label{tab:oddeven-params}
\begin{center}
\small
\begin{tabular}{lp{9.2cm}}
\toprule
 & Odd--Even \\
\midrule
Hidden state $s^\ast$ & one integer drawn uniformly from $\{1,\dots,50\}$ at reset; fixed for the episode \\
Episode length & 30 steps, with a prediction at every step \\
Observation $o_t$ & one integer per step, drawn from a discretized Gaussian centered on $s^\ast$ with $\sigma=\sqrt{5}+1\approx3.24$ and restricted to the integers in $[1,50]$ of the same parity as $s^\ast$; one observation at reset and one after every action, so the prediction at step $t$ uses the first $t$ observations \\
Action & discrete, 50 choices: a prediction of $s^\ast$ \\
Reward & $+1$ for an exact match, $0$ otherwise; no shaping \\
Belief $b_t$ & the exact posterior over the 50 states, one particle per integer with the posterior as its weight, from a uniform prior; every opposite-parity state has weight exactly $0$ from the first observation on, and the mass on the remaining 25 states sharpens with each observation, $\max_s b_t(s)>0.9$ by step 21 \\
Belief frame & $(s-25.5)/24.5\in[-1,1]$; 50 particles \\
Policy input besides the belief & the normalized step index $t/30$ \\
Termination & timeout at 30 steps \\
Evaluation metric & exact-match rate over the steady-state steps 22--30; the transient steps 1--21 are excluded \\
Reference policies & steady-state exact-match rate of the posterior mode $0.88$, of repeating the previous observation $0.26$, of a uniform random guess $0.02$ \\
\bottomrule
\end{tabular}
\end{center}
\end{table}

%% file: tables/results/table_frame_stacking.tex
\begin{table}[h]
\caption{Frame-stacking results, five PPO seeds, the evaluation protocol of
Section~\ref{sec:experiments:protocol}. Beacon Sweep: fraction of beacons
reached / all-five rate. Faint and Diffuse Beacon: correct rate, and in
parentheses the correct rate among episodes that end in a contact (chance
$0.2$).}
\label{tab:framestack-results}
\begin{center}
\small
\begin{tabular}{llp{4.6cm}l}
\toprule
Task & $k$ & Score & Other $k$ \\
\midrule
CF & 160 & 0.80 (three seeds solve every episode; two commit to one end) & -- \\
AT & 8 & 0.017 & -- \\
LAT & 8 & 0.000 & -- \\
HAT & 8 & 0.730 & -- \\
OE & 30 & 0.260 & $k=8$: 0.242 \\
BS & 60 & 0.24 / 0.003 & $k=20$: 0.24 / 0.001; $k=1$: 0.19 / 0.000 \\
FB & 60 & 0.116 (0.185) & $k=20$: 0.135 (0.200) \\
DB & 60 & 0.136 (0.191) & $k=20$: 0.140 (0.210) \\
\bottomrule
\end{tabular}
\end{center}
\end{table}

%% file: tables/results/table_frame_stacking_family.tex
\begin{table}[H]
\caption{The frame-stacking baseline's input per task family.}
\label{tab:framestack-input}
\begin{center}
\small
\begin{tabular}{lp{8.2cm}cc}
\toprule
Task & Frame & $k$ & Input width \\
\midrule
CF & position, velocity, priest reading (the native observation) & 160 & 480 \\
AT, LAT & the 31-d MuJoCo observation, with the target's coordinates unmasked while inside the visibility radius and zero outside it & 8 & 248 \\
HAT & as above, plus a constant 6-vector: the two active dens (scaled by the arena half-width) and their prior masses, the filter's initial belief & 8 & 254 \\
OE & the observation and the time index, $(t/30,\;o_t/50)$ & 30 & 60 \\
BS, FB, DB & agent position, then five slots of (bit, $\Delta x$, $\Delta y$): this step's anonymous pings in random slot order, agent-relative and scaled; empty slots zero & 60 & 1020 \\
\bottomrule
\end{tabular}
\end{center}
\end{table}

%% file: figures/ablate.tex
\begin{figure}[H]
\centering
\includegraphics[width=0.6\linewidth]{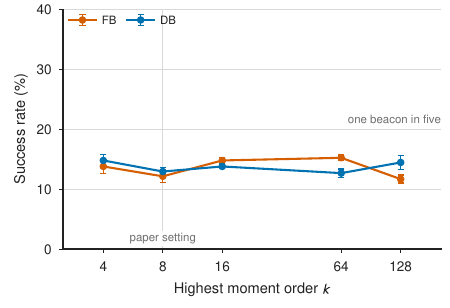}
\caption{Sensitivity of the moment encoder to the highest moment order $k$
on Faint Beacon (FB) and Diffuse Beacon (DB). Success rate is the fraction of
300 evaluation episodes in which the agent reaches the target beacon; points
show the mean $\pm$ standard error over three PPO seeds (five at the main-text setting). The vertical line
marks the setting used in the main text, and the dashed line marks $20\%$, the
rate of picking one of the five beacons uniformly at random.}
\label{fig:ablation-moment-order}
\end{figure}

\begin{figure}[H]
\centering
\includegraphics[width=0.6\linewidth]{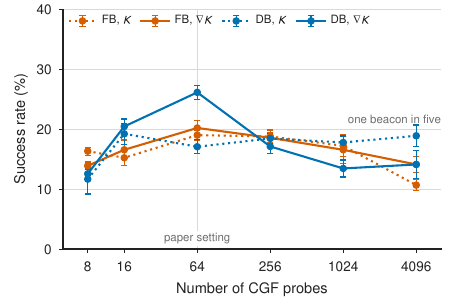}
\caption{Sensitivity of the CGF encoder to the number of probes on Faint
Beacon (FB) and Diffuse Beacon (DB), using CGF values ($K$, dotted) or
gradients ($\nabla K$, solid) as features, with learned probe locations. Success rate, error bars, and
reference lines as in Figure~\ref{fig:ablation-moment-order}.}
\label{fig:ablation-cgf-probes}
\end{figure}